\documentclass[preprint,12pt]{elsarticle}

\journal{Green Technologies and Sustainability}

\usepackage{amsmath,amssymb,amsfonts}
\usepackage{amsthm}

\usepackage{graphicx}
\usepackage{booktabs}
\usepackage{array}
\usepackage{tabularx}

\usepackage{subcaption}

\usepackage{textcomp}
\usepackage{url}

\biboptions{sort&compress}

\begin{document}

\begin{frontmatter}



\title{Structural Gating and Effect-aligned Lag-resolved Temporal Causal Discovery Framework with Application to Heat–Pollution Extremes}
\author[aff1]{Rui Chen}
\author[aff1,aff2]{Jinsong Wu\corref{cor1}}
\cortext[cor1]{Corresponding author:}
\ead{wujs@ieee.org}

\affiliation[aff1]{organization={School of Artificial Intelligence},
                   addressline={Guilin University of Electronic Technology},
                   city={Guilin},
                   state={Guangxi},
                   postcode={541004},
                   country={China}}

\affiliation[aff2]{organization={Department of Electrical Engineering},
                   addressline={University of Chile},
                   city={Santiago},
                   state={Santiago Metropolitan Region},
                   postcode;={8370451},
                   country={Chile}}

\begin{abstract}
This study proposes Structural Gating and Effect-aligned Discovery for
Temporal Causal Discovery (SGED-TCD), a novel and general framework for
lag-resolved causal discovery in complex multivariate time series.
SGED-TCD combines explicit structural gating, stability-oriented learning,
perturbation-effect alignment, and unified graph extraction to improve the
interpretability, robustness, and functional consistency of inferred causal
graphs.

To evaluate its effectiveness in a representative real-world setting, we apply
SGED-TCD to teleconnection-driven compound heatwave--air-pollution extremes
in eastern and northern China. Using large-scale climate indices, regional
circulation and boundary-layer variables, and compound extreme indicators, the
framework reconstructs weighted causal networks with explicit dominant lags
and relative causal importance. The inferred networks reveal clear regional
and seasonal heterogeneity: warm-season extremes in Eastern China are mainly
linked to low-latitude oceanic variability through circulation, radiation,
and ventilation pathways, whereas cold-season extremes in Northern China are
more strongly governed by high-latitude circulation variability associated
with boundary-layer suppression and persistent stagnation.

These results show that SGED-TCD can recover physically interpretable, hierarchical, and lag-resolved causal pathways in a challenging climate--environment system. More broadly, the proposed framework is not restricted to the present application and provides a general basis for temporal causal discovery in other complex domains.
\end{abstract}




\end{frontmatter}



\section{Introduction}
In recent decades, \emph{concurrent heatwave--air-pollution extremes} have intensified in both frequency and magnitude across eastern and northern China, posing significant threats to public health, energy infrastructure, and socio-economic stability. Compared with isolated heat or pollution events, such compound extremes manifest pronounced nonlinear amplification effects. Specifically, elevated temperatures intensify near-surface ozone (O$_3$) formation and accumulation of particulate matter (PM$_2.5$), leading to disproportionately severe environmental and health impacts \cite{xiao2022amplified,cui2025future,pu2017enhanced}. Consequently, elucidating the physical drivers of these concurrent extremes is of critical importance for robust climate risk assessment and adaptation planning.
Accumulated evidences suggest that heatwave-air-pollution compound events are not merely governed by local emissions or short-term meteorological variability but are strongly modulated by \emph{large-scale atmosphere--ocean circulation anomalies} via remote teleconnection processes \cite{wu2015has,jiang2021impact,zhang2019possible}. Climate modes—including the El~Ni~no--Southern Oscillation (ENSO), tropical Indian and Pacific sea surface temperature anomalies, and mid- to high-latitude circulation patterns—can excite planetary-scale Rossby wave trains, systematically altering the background circulation over East Asia. These large-scale adjustments regulate key regional meteorological conditions, such as subsidence strength, boundary-layer stability, radiative forcing, and ventilation efficiency, thereby fostering conditions conducive to the co-occurrence of extreme heat and severe air pollution.
During the warm season, concurrent heatwave--ozone extremes over eastern China have been closely linked to anomalies of the \emph{Western Pacific Subtropical High (WPSH)}. A strengthened or westward-extended WPSH typically induces persistent subsidence, enhanced solar radiation, and weak near-surface winds. These conditions not only increase the likelihood of extreme heat, but also facilitate ozone accumulation by suppressing pollutant dispersion and accelerating photochemical reactions \cite{wu2015has,jiang2021impact}. In contrast, during the cold season, compound pollution extremes in northern China—often dominated by PM$_2.5$ - are more profoundly influenced by \emph{mid--high latitude teleconnection patterns}, including the Arctic Oscillation (AO), the North Atlantic Oscillation (NAO) and North Atlantic sea surface temperature anomalies. These large-scale drivers modulate the intensity of the East Asian winter monsoon and the frequency of cold-air intrusions, thus controlling atmospheric stagnation and pollutant accumulation \cite{lu2021impact,zhang2019seesaw}.
Despite these advances, several critical limitations persist. Most existing studies have predominantly relied on correlation-based analyzes, composite diagnostics, or regression approaches, which are inadequate to disentangle direct causal effects from indirect modulation or common forcing. Furthermore, teleconnection influences are inherently \emph{time-lagged, nonlinear, and nonstationary}, and the bidirectional coupling between heat and air pollution further complicates the causal structure of compound extremes. These challenges require a methodological framework capable of explicitly resolving time-dependent causal relationships within a multivariate setting \cite{granger1969investigating,runge2019detecting}.

To address this gap, this study proposes Structural Gating and Effect-aligned
Discovery for Temporal Causal Discovery (SGED-TCD). This framework is developed 
to improve the recovery of lag-resolved causal structure in complex multivariate 
systems where dependencies are nonlinear, temporally delayed, and strongly autocorrelated. SGED-TCD is designed as an independent framework for identifying
lag-specific, directed, and interpretable causal structure from time series.
By replacing indirect attention-based structural inference with explicit
structural gates, stability constraints, and perturbation-effect alignment,
the framework provides a more robust and interpretable basis for temporal
causal graph discovery. These methodological properties make SGED-TCD
well suited to Earth system and climate--environment applications, in which
remote drivers often influence target extremes through hierarchical and
multi-step pathways.

\section{Data and Methodology}
\subsection{SGED-TCD-Based Causal Discovery Framework}

This study proposes Structural Gating and Effect-aligned Discovery for Temporal Causal Discovery (SGED-TCD) as the core methodological framework for identifying lag-specific, directed, and interpretable causal structure from time series. The framework is designed to identify lag-specific and interpretable causal pathways from multivariate time series by explicitly parameterizing candidate causal links, rather than inferring them indirectly from attention weights.

The methodological premise of SGED-TCD is that reliable temporal causal
discovery should satisfy four requirements simultaneously: it should
(i) represent candidate source--target--lag relations explicitly,
(ii) maintain structural consistency under small perturbations of the input,
(iii) distinguish causally meaningful relations from prediction-equivalent but
structurally spurious alternatives, and (iv) align inferred structural
importance with actual functional influence on prediction. To achieve these
goals, SGED-TCD adopts a backbone-decoupled architecture consisting of a
variable-level temporal encoder, a conditional structural gating tensor, a
lag-aware target-node aggregator, and a prediction head.

\subsubsection{General suitability of SGED-TCD for temporal causal discovery}

Temporal causal discovery in real-world systems often faces several recurring
difficulties: nonlinear interactions, multistep dependence chains, strong
autocorrelation, and ambiguity between predictive relevance and structural
importance. SGED-TCD is designed to address these challenges through the
following properties.

\paragraph{Explicit lag-resolved structural representation}
Instead of relying on post hoc interpretation of attention distributions,
SGED-TCD introduces explicit structural gates $Z_{i,j,\tau}$ for candidate
source--target--lag relations. This makes graph structure a first-class
learnable object and provides direct lag-resolved causal interpretation.

\paragraph{Capacity for nonlinear and interacting pathways}
The variable-level temporal encoder and lag-aware aggregation module enable
SGED-TCD to capture nonlinear mappings and interaction effects, making the
framework applicable to systems in which causal influences are mediated
through complex intermediate processes rather than simple pairwise effects.

\paragraph{Improved structural stability}
In many temporal datasets, multiple predictive solutions may achieve similar
forecasting performance while implying different graph structures. SGED-TCD
addresses this issue by enforcing consistency of inferred structure across
structure-preserving perturbation views, thereby improving reproducibility and
reducing sensitivity to minor input changes.

\paragraph{Alignment between structural importance and functional effect}
A key feature of SGED-TCD is that structural prominence alone is not treated
as sufficient evidence for causal relevance. Instead, the framework evaluates
whether candidate links produce measurable predictive impact through
perturbation-based effect estimation, thereby improving the functional
credibility of the inferred graph.

\subsubsection{Model input, lag window, and training objective}

Let $X(t)\in\mathbb{R}^N$ denote the standardized multivariate time-series
vector. For each target node $j$, SGED-TCD learns a one-step-ahead temporal
predictor from the lagged multivariate history:
\begin{equation}
\hat{X}_j(t)=f_j\big(X(t-1),X(t-2),\dots,X(t-L)\big),
\end{equation}
where $L$ denotes the maximum lag window.

Unlike attention-based formulations, SGED-TCD decomposes temporal causal
discovery into two coordinated components: temporal feature encoding and
explicit structural gating. First, a variable-level encoder transforms the
lagged input sequence into hidden representations. Second, a conditional
structural gating tensor determines how source representations at different
lags contribute to the prediction of each target variable. The gate logits are
defined as
\begin{equation}
\mathrm{logits}_{i,j,\tau}
=
\mathrm{base\_gate\_logits}_{i,j,\tau}
+
\mathrm{dynamic\_logits}_{i,j,\tau},
\end{equation}
where the first term represents a global learnable structural tendency and
the second term is conditioned on source--target interaction features derived
from the encoded sequence.

For target variable $j$, the lag-aware aggregated message is written as
\begin{equation}
m_j^t
=
\sum_{i=1}^{N}\sum_{\tau=1}^{L}
Z_{i,j,\tau}\,\phi\!\left(h_i^{t-\tau},h_j^t\right),
\end{equation}
where $h_i^{t-\tau}$ and $h_j^t$ denote encoded source and target
representations, respectively, and $\phi(\cdot)$ is a learnable
transformation. The final prediction is then obtained by combining the
aggregated message with the target representation:
\begin{equation}
\hat{x}_j^t=g_j(m_j^t,h_j^t).
\end{equation}

The overall training objective of SGED-TCD consists of four components:
\begin{equation}
\mathcal{L}_{\mathrm{total}}
=
\mathcal{L}_{\mathrm{pred}}
+
\lambda_{\mathrm{sparse}}\mathcal{L}_{\mathrm{sparse}}
+
\lambda_{\mathrm{stable}}\mathcal{L}_{\mathrm{stable}}
+
\lambda_{\mathrm{abl}}\mathcal{L}_{\mathrm{abl}}.
\end{equation}
Here, $\mathcal{L}_{\mathrm{pred}}$ is the prediction loss,
$\mathcal{L}_{\mathrm{sparse}}$ encourages sparse graph discovery,
$\mathcal{L}_{\mathrm{stable}}$ enforces structural consistency across
perturbation views, and $\mathcal{L}_{\mathrm{abl}}$ aligns learned
structural gates with perturbation-induced predictive effects. This
multi-objective design enables the framework to balance predictive fidelity,
sparsity, stability, and causal interpretability.

\subsubsection{From structural gates to candidate causal edges}

After training, SGED-TCD yields two lag-specific quantities for each ordered
source--target pair: the learned structural gate $Z_{i,j,\tau}$ and the
perturbation-based effect estimate $\delta_{i,j,\tau}$. For each candidate
triplet $(i,j,\tau)$, a composite edge score is constructed as
\begin{equation}
S_{i,j,\tau}
=
\alpha\,\bar{Z}_{i,j,\tau}
+
\beta\,\bar{\delta}_{i,j,\tau},
\end{equation}
where $\bar{Z}_{i,j,\tau}$ and $\bar{\delta}_{i,j,\tau}$ denote normalized
gate and effect quantities, respectively, and $\alpha$ and $\beta$ are
weighting coefficients.

For each ordered variable pair $(i\rightarrow j)$, the dominant lag is
identified as
\begin{equation}
\tau^{*}_{i\rightarrow j}
=
\arg\max_{\tau\in[1,L]} S_{i,j,\tau},
\end{equation}
and the corresponding lag-aggregated causal strength is obtained by
\begin{equation}
A_{i,j}
=
\max_{\tau\in[1,L]} S_{i,j,\tau}.
\end{equation}

Edges are retained only if $A_{i,j}$ exceeds a data-driven graph-selection
threshold. In this way, SGED-TCD converts explicit structural gates and
perturbation-aligned evidence into a sparse and interpretable lag-aware
causal network. This graph extraction strategy is general and can be used in
different temporal domains, irrespective of the specific semantics of the
variables.

\subsubsection{Causal validation and robustness checks}

Although SGED-TCD improves structural identifiability by introducing explicit
gates, inferred edges remain subject to uncertainty arising from common
drivers, autocorrelation, finite sample size, and application-specific noise.
The framework therefore incorporates several robustness procedures to improve
the credibility of the final causal graph.

\paragraph{Ablation-based validation.}
For each candidate edge $(i\rightarrow j)$, the predictive performance of
$f_j$ is re-evaluated after ablating the lagged history of source variable $i$
around the dominant lag $\tau^{*}_{i\rightarrow j}$. A substantial
performance degradation supports the interpretation that the source provides
non-redundant predictive information.

\paragraph{Stability under perturbed views.}
SGED-TCD evaluates whether the inferred structure remains consistent across
structure-preserving perturbation views generated from the same time series.
Links that persist under such perturbations are regarded as more reliable than
links that fluctuate strongly under minor input changes.

\paragraph{Lag-sensitivity stability.}
The identified dominant lag $\tau^{*}_{i\rightarrow j}$ is further examined
under moderate changes in the maximum lag window $L$ and under resampling
procedures, so that retained relations reflect stable temporal scales rather
than artifacts of a particular configuration.

\paragraph{Cross-method plausibility screening.}
Where appropriate, high-confidence SGED-TCD links can be compared with
well-established theoretical, empirical, or domain-specific constraints. This
step does not require exact agreement across methods, but provides an
additional plausibility check for interpreting the discovered pathways.

Overall, SGED-TCD is intended as a general temporal causal discovery
framework. The teleconnection-driven compound-extreme problem studied here
serves as a representative application scenario for evaluating its ability to
recover lag-resolved, interpretable, and physically meaningful causal
structure in a challenging real-world setting.

\subsection{Study Regions and Target Variables}

\begin{table}[!t]
\centering
\caption{Definitions for Each Considered Variable}
\label{tab:variables}
\footnotesize
\setlength{\tabcolsep}{4pt}
\renewcommand{\arraystretch}{1.05}
\begin{tabularx}{\textwidth}{l >{\arraybackslash}X >{\arraybackslash}X}
\hline
\textbf{Abbreviation} & \textbf{Variable Description} & \textbf{Spatial Domain (Latitude--Longitude)} \\
\hline
Ni\~no3.4 & ENSO index representing SST anomalies in the central--eastern tropical Pacific & 5$^\circ$S--5$^\circ$N, 170$^\circ$W--120$^\circ$W \\
IO\_SST & Indian Ocean sea surface temperature anomaly index & 20$^\circ$S--20$^\circ$N, 40$^\circ$E--100$^\circ$E \\
WP\_SST & Western Pacific warm pool sea surface temperature anomaly index & 0$^\circ$--20$^\circ$N, 120$^\circ$E--160$^\circ$E \\
NA\_SST & North Atlantic sea surface temperature anomaly index & 0$^\circ$--60$^\circ$N, 80$^\circ$W--0$^\circ$ \\
AO & Arctic Oscillation index & 20$^\circ$N--90$^\circ$N \\
NAO & North Atlantic Oscillation index & 20$^\circ$N--80$^\circ$N, 90$^\circ$W--40$^\circ$E \\
Z500 & 500-hPa geopotential height anomaly & 20$^\circ$--60$^\circ$N, 100$^\circ$E--140$^\circ$E \\
T2m & Near-surface air temperature at 2 m & Eastern China / Northern China \\
WS10 & 10-m wind speed & Eastern China / Northern China \\
PBLH & Planetary boundary layer height & Eastern China / Northern China \\
RH & Near-surface relative humidity & Eastern China / Northern China \\
SW$\downarrow$ & Surface downward shortwave radiation & Eastern China / Northern China \\
HW\_Int & Heatwave intensity index & Eastern China / Northern China \\
HW\_Dur & Heatwave duration (days) & Eastern China / Northern China \\
O$_3$ & Surface ozone concentration (warm season) & Eastern China (25$^\circ$--35$^\circ$N, 110$^\circ$--125$^\circ$E) \\
PM$_{2.5}$ & Fine particulate matter concentration (cold season) & Northern China (35$^\circ$--42$^\circ$N, 112$^\circ$--120$^\circ$E) \\
\hline
\end{tabularx}
\end{table}

This study targets two climatically and environmentally distinct regions in China: \emph{Eastern China (EC)} and \emph{Northern China (NC)}, where concurrent heatwave--air-pollution extremes exhibit pronounced regional heterogeneity. Here, the two study domains are defined from a process-oriented perspective rather than by administrative boundaries. Specifically, EC is selected to represent the core warm-season hotspot of heatwave--ozone co-occurrence, whereas NC is selected to represent the core cold-season hotspot of heat-related stagnant conditions accompanied by PM$_{2.5}$ accumulation. The regional boundaries are therefore chosen to capture the dominant spatial footprints of these two compound-risk regimes and to maintain consistency with the pollutant target domains listed in Table~\ref{tab:variables}. In EC, warm-season compound extremes manifest primarily as heatwave--ozone (O$_3$) events, whereas in NC, cold-season compound risks are dominated by fine particulate matter (PM$_{2.5}$) co-occurring with anomalously warm and stagnant conditions.

To characterize thermal extremes, two complementary heatwave indices are adopted: heatwave intensity (HW\_Int) and heatwave duration (HW\_Dur). The extremes of air pollution are represented by the concentrations of surface O$_3$ during the warm season in the EC and the concentrations of PM$_{2.5}$ during the cold season in NC. All target variables are aggregated to a monthly resolution to facilitate the investigation of teleconnection-driven variability and delayed responses.
\subsection{Datasets and Variables}
We constructed a comprehensive monthly multivariate dataset to investigate whether large-scale atmosphere--ocean teleconnections exert lagged impacts on compound heatwave-air-pollution extremes over Eastern China (EC) and Northern China (NC). All variables and spatial domains are detailed in Table~\ref{tab:variables}. The data sources, preprocessing steps, regional/seasonal definitions, and derived metrics are summarized below.

\subsubsection{Teleconnections and SST-based indices}
The monthly Ni\~no3.4 index is sourced from NOAA PSL \cite{NOAA_PSL_NINO34}. The AO and NAO indices are obtained from NOAA CPC teleconnections products \cite{NOAA_CPC_Teleconnections}. To consistently represent basin-scale SST variability, IO\_SST, WP\_SST, and NA\_SST are derived from NOAA ERSSTv5 monthly SST data \cite{huang2017extended} via latitude-weighted box averaging, with anomalies calculated relative to a fixed climatology (e.g., 1991--2020).

\subsubsection{Regional circulation and meteorological mediators}
Regional circulation and boundary-layer conditions are characterized using the ERA5 reanalysis \cite{hersbach2020era5}, specifically focusing on Z500 anomalies and PBLH, along with near-surface meteorological variables as required. For descriptive mediator analyses and visualizations, monthly regional aggregates of T2m, WS10, RH, and SW$\downarrow$ were retrieved via the NASA POWER API \cite{NASA_POWER_API}. All gridded variables are aggregated over EC and NC using latitude-weighted area averaging. Specifically, for all grid cells located within the target domain, the regional monthly mean is computed as
\begin{equation}
\bar{X}(t)=\frac{\sum_{i,j} X_{i,j}(t)\cos(\phi_i)}{\sum_{i,j}\cos(\phi_i)},
\end{equation}
where $X_{i,j}(t)$ denotes the value at grid cell $(i,j)$ in month $t$, and $\phi_i$ is the latitude of the grid-cell center. The weight $\cos(\phi_i)$ is used to account for the decrease in grid-cell area with latitude. After spatial aggregation, monthly anomalies are derived by removing the corresponding climatological monthly mean over the reference period.

\subsubsection{Air pollution and heatwave metrics}
Monthly near-surface O$_3$ and PM$_{2.5}$ data are extracted from the global reanalysis of CAMS (EAC4) and spatially averaged over the corresponding domains \cite{acp-19-3515-2019}. The duration and intensity of the heat waves (HW\_Dur and HW\_Int) are derived from the daily ERA5 T2m \cite{hersbach2020era5} using a percentile-based threshold subject to a minimum-duration constraint; the daily events are then aggregated into monthly counts (HW\_Dur) and intensity based on exceedance (HW\_Int). Reflecting distinct regional atmospheric chemistry regimes, the analysis emphasizes O$_3$ during the warm season and PM$_{2.5}$ during the cold season.

\subsection{Exploratory Analysis of Compound Extremes and Teleconnections}
\begin{figure}[t]
  \centering
  \includegraphics[width=1\textwidth]{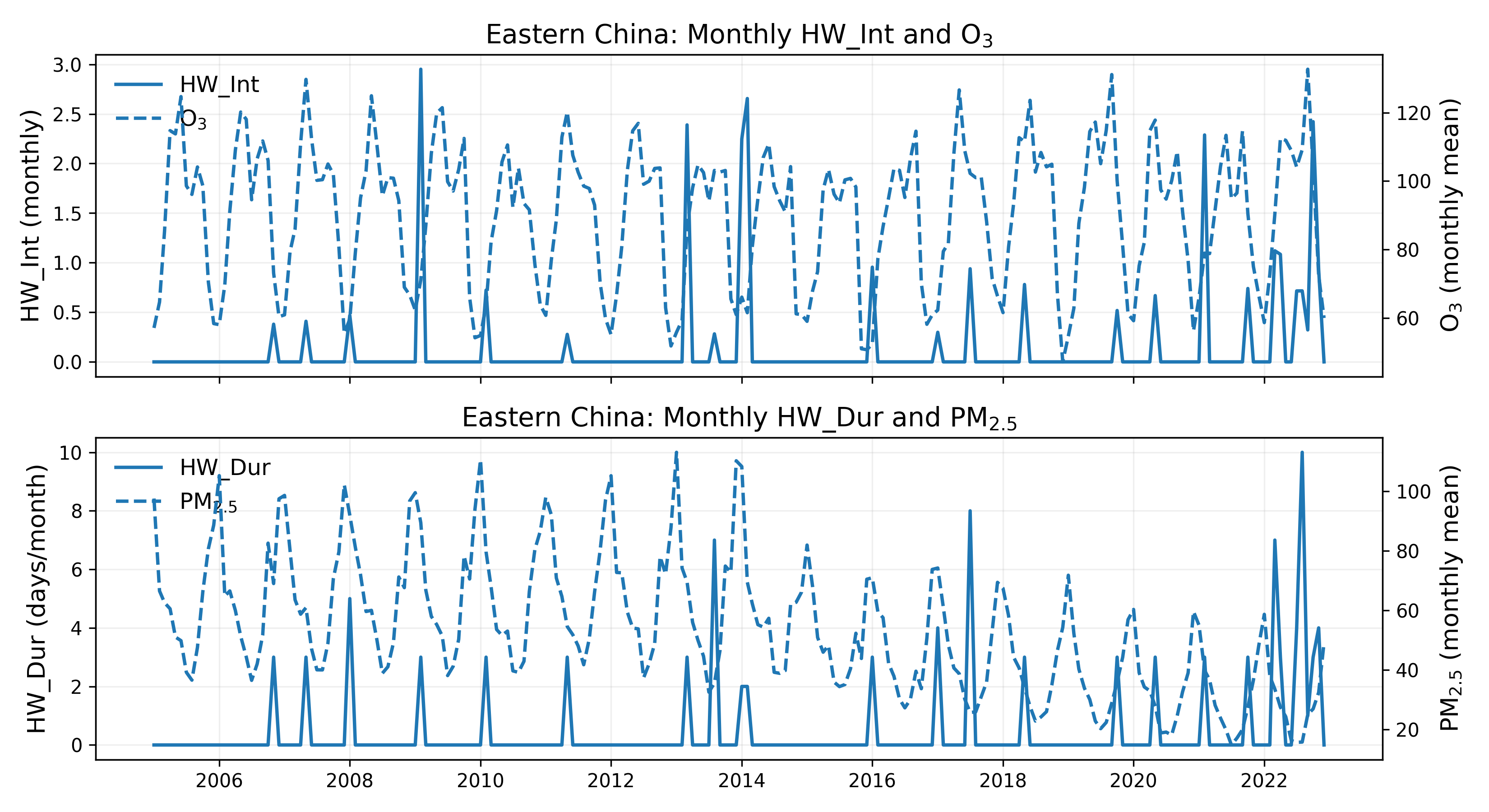}
  \includegraphics[width=1\textwidth]{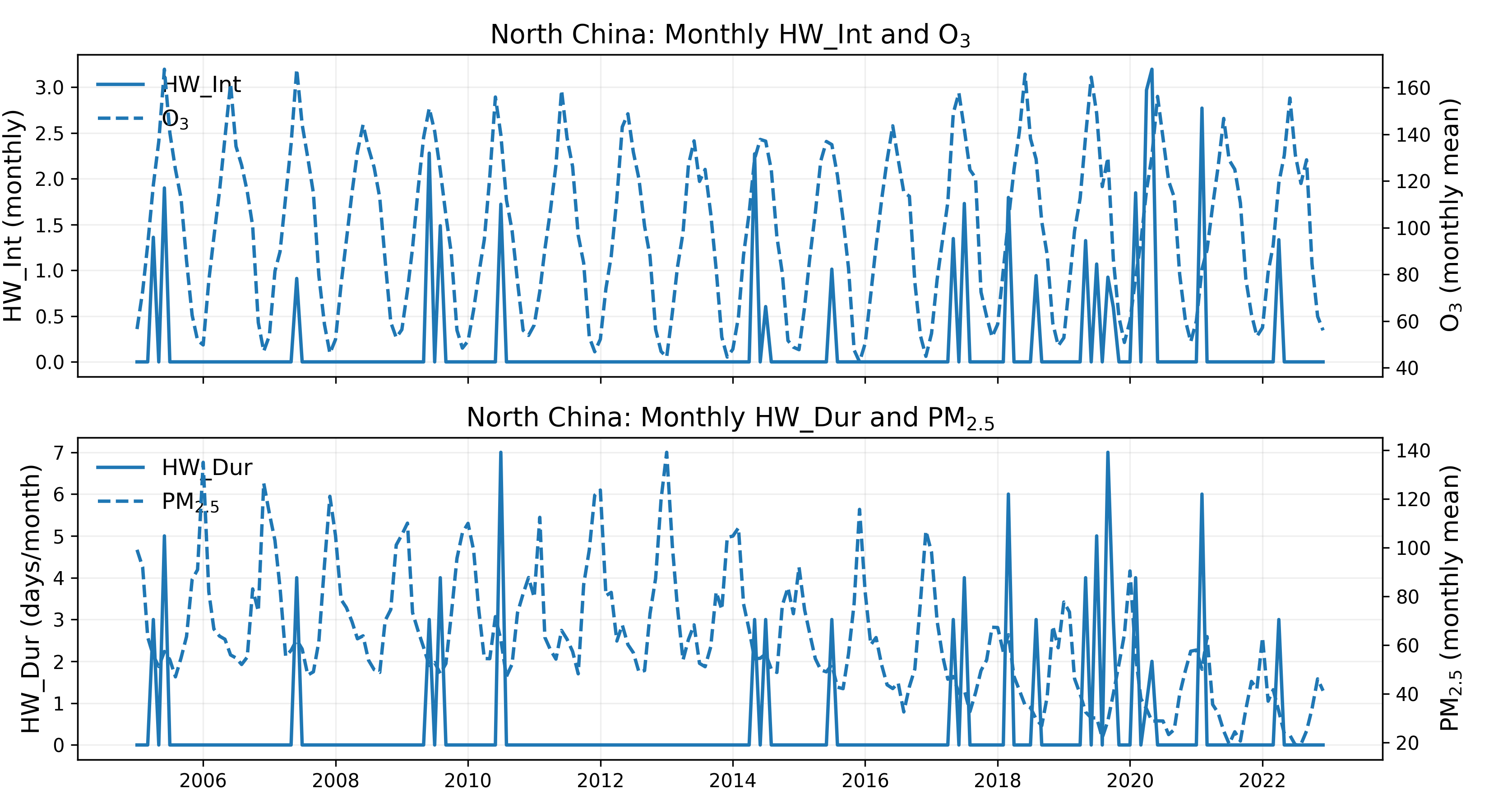}
  \caption{Monthly time series of heatwave indices and air pollution variables over Eastern China (EC) and Northern China (NC).}
  \label{fig:monthly_ts}
\end{figure}

Fig.~\ref{fig:monthly_ts} shows the monthly evolution of the heatwave indices and air pollutant concentrations in the EC and NC. In EC, a pronounced warm-season synchronization between HW\_Int and O$_3$ is evident, with peak ozone concentrations systematically aligning with periods of enhanced heatwave intensity. 
Conversely, PM$_{2.5}$ over NC exhibits a strong cold-season cycle, while HW\_Dur occasionally intensifies under anomalously warm and stagnant cold-season conditions, underscoring the seasonal asymmetry of compound risks between the two regions.

\begin{figure}[t]
  \centering
  \includegraphics[width=1\textwidth]{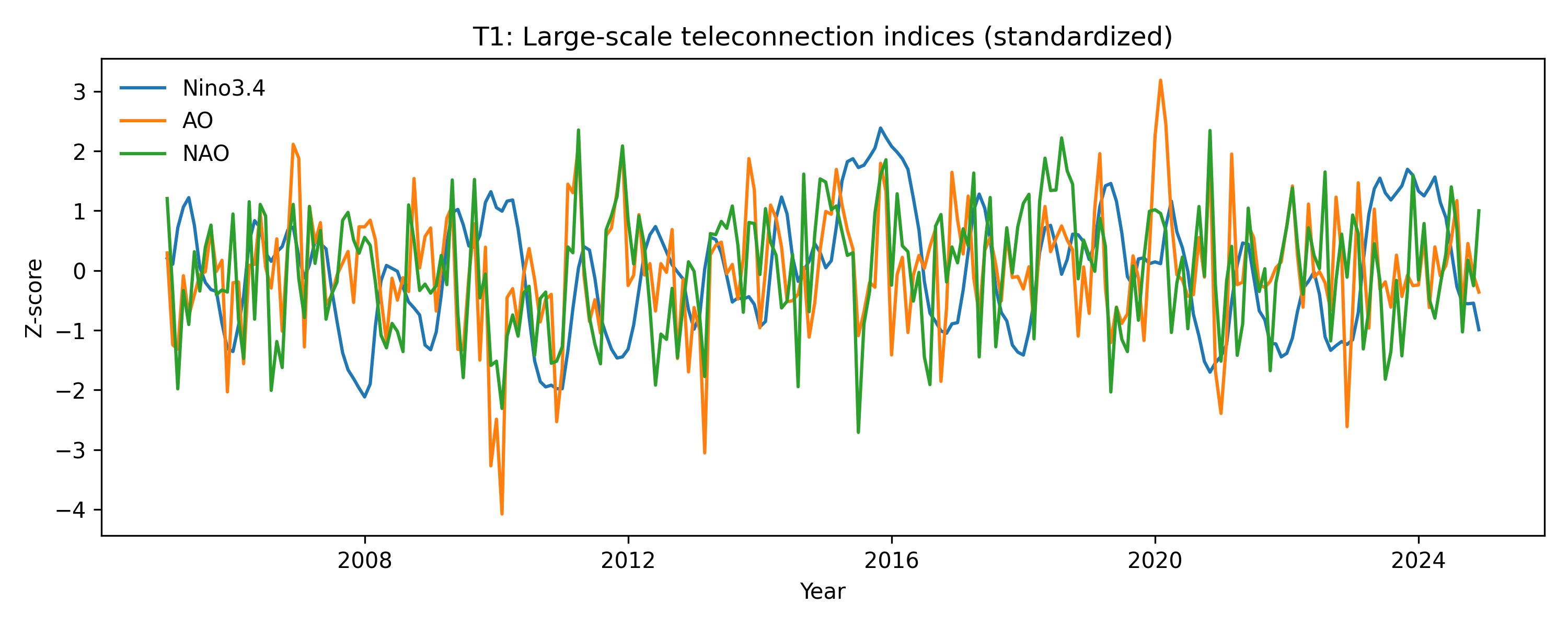}
  \caption{Standardized time series of major large-scale teleconnection indices.}
  \label{fig:tele_ts}
\end{figure}

\begin{figure}[t]
  \centering
  \includegraphics[width=1\textwidth]{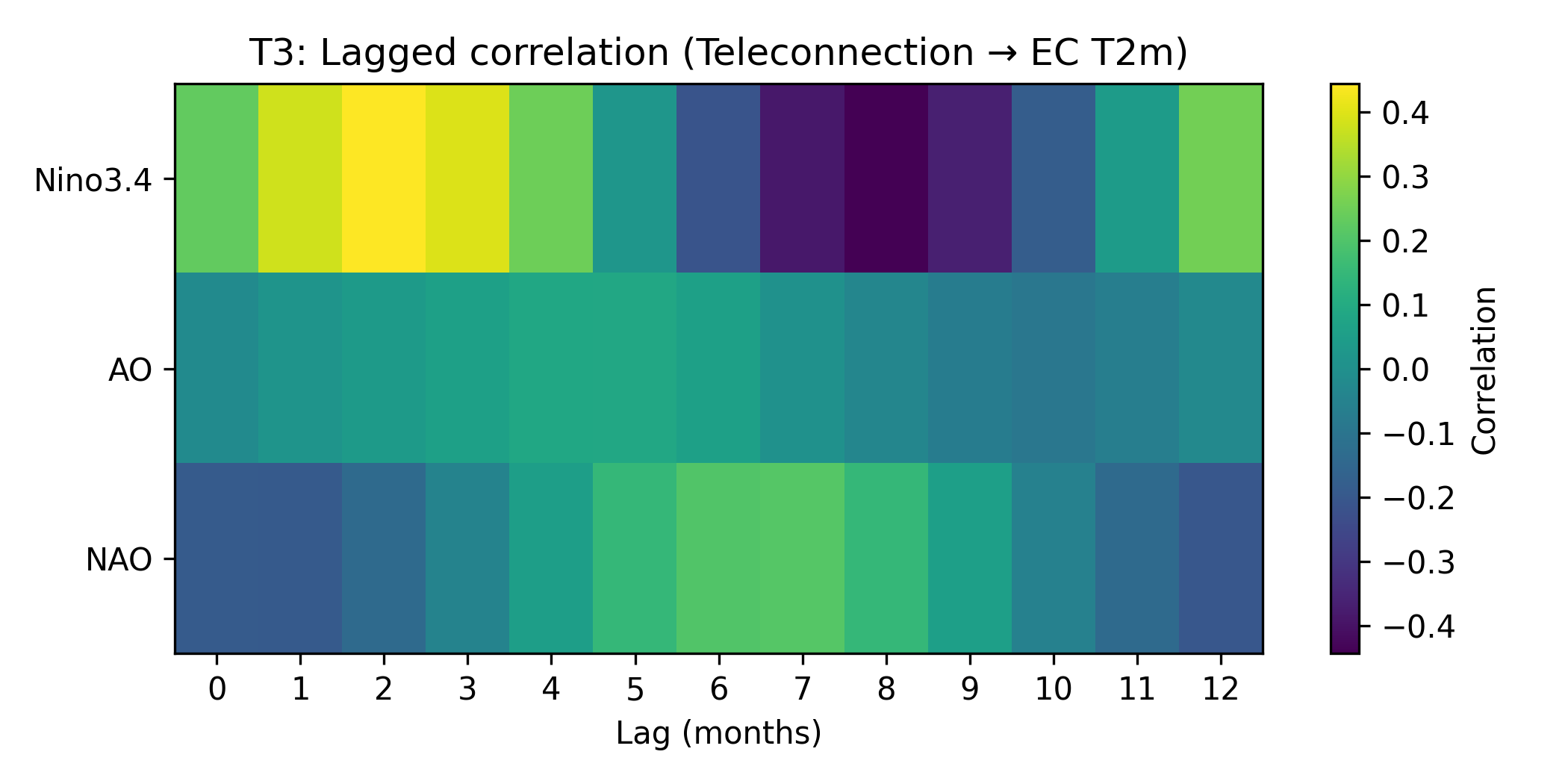}
  \caption{Lagged correlations between teleconnection indices and EC near-surface temperature (T2m).}
  \label{fig:lagcorr}
\end{figure}

Large-scale teleconnection variability is presented in Fig.~\ref{fig:tele_ts}, where standardized Ni\~no3.4, AO, and NAO indices exhibit substantial interannual fluctuations, suggesting potential delayed and nonlinear influences on regional climate conditions. To provide a preliminary assessment of delayed responses, Fig.~\ref{fig:lagcorr} displays lagged correlations between teleconnection indices and EC near-surface temperature. In particular, Ni\~no3.4 exhibits a distinct positive correlation at lags of 1--4 months, while NAO shows peak correlations at longer lags, highlighting the need to explicitly model time-delayed causal effects.

\begin{figure*}[t]
  \centering
  \includegraphics[width=1\textwidth]{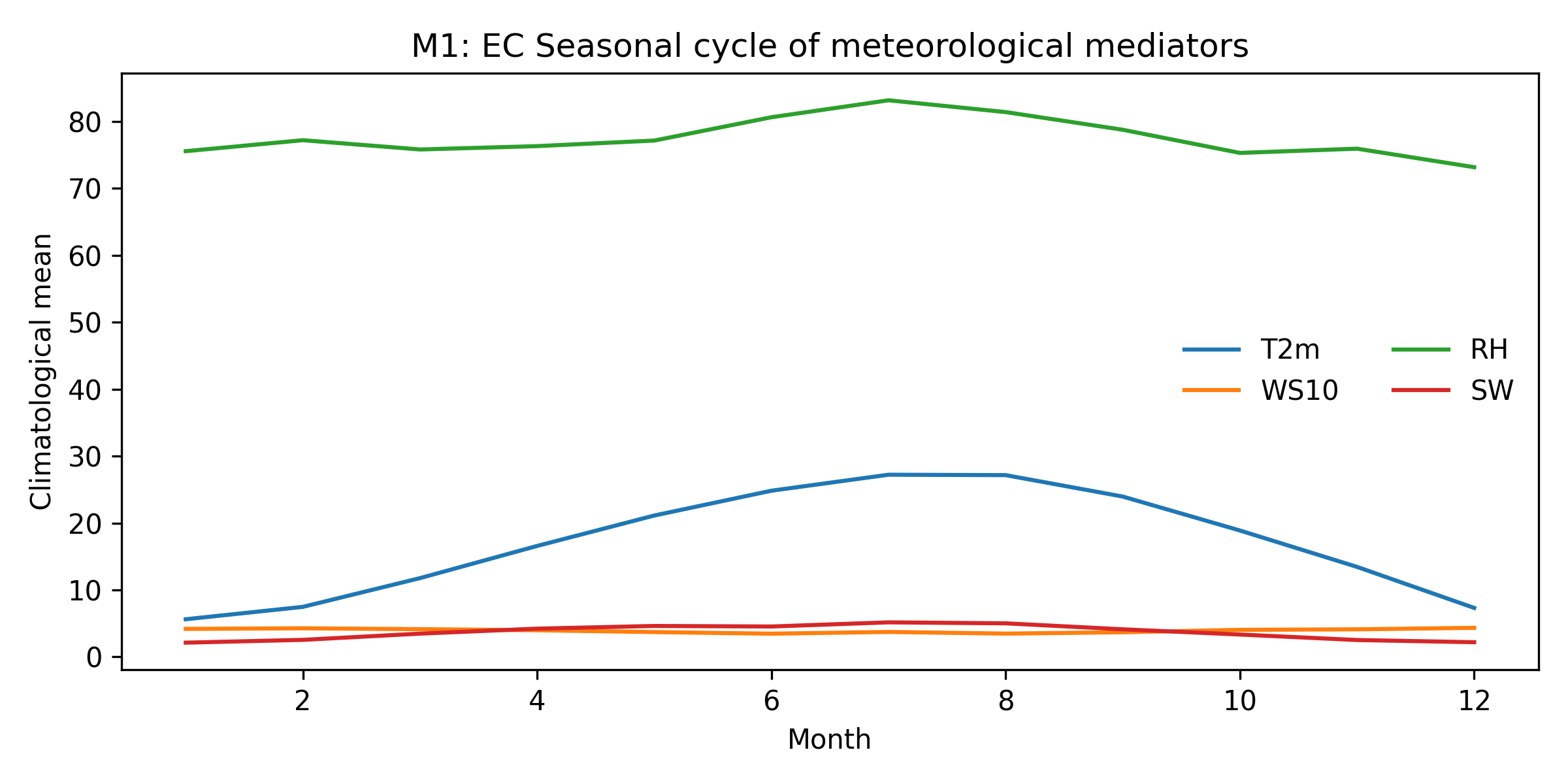}
  \includegraphics[width=1\textwidth]{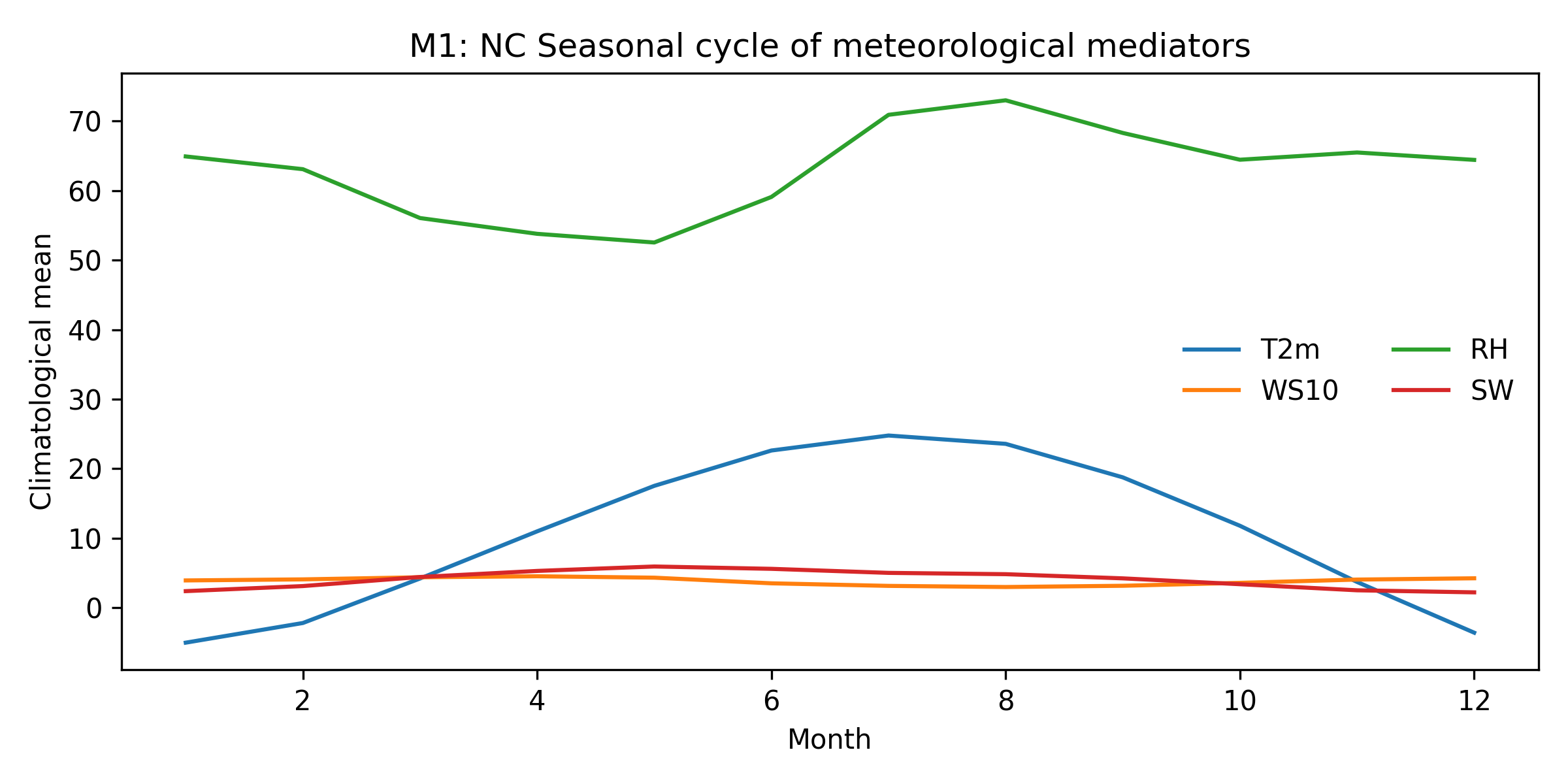}
  \caption{Seasonal cycles of key meteorological mediators over EC and NC.}
  \label{fig:seasonal}
\end{figure*}

The seasonal climatologies of the meteorological mediators Fig.~\ref{fig:seasonal} further indicate that compound extremes are preferentially associated with high-temperature, high-radiation, weak-wind, and relatively humid conditions, consistent with established physical mechanisms governing heatwave persistence and accumulation of pollution.

\begin{figure}[t]
  \centering
  \includegraphics[width=1\textwidth]{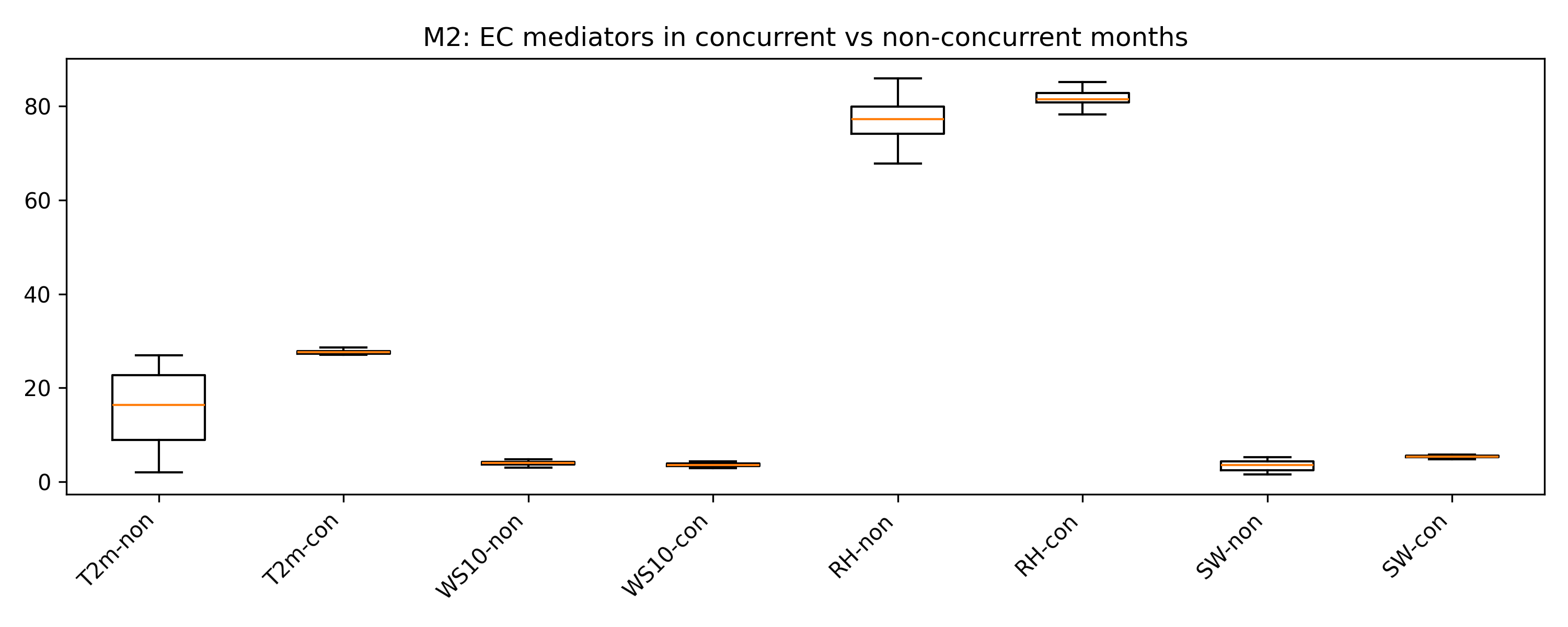}
  \caption{Comparison of meteorological mediators during concurrent and non-concurrent months in EC.}
  \label{fig:box}
\end{figure}

Finally, Fig.~\ref{fig:box} contrasts meteorological conditions during concurrent and non-concurrent months in EC, revealing systematically higher temperature, humidity, and shortwave radiation during compound extreme periods, thereby motivating a causal mediation perspective.

\section{Experimental Design and Causal Network Construction}

\subsection{Experimental Setup}

Causal discovery experiments are conducted independently for Eastern China (warm season) and Northern China (cold season), following the region--season stratification introduced in Section 2. Monthly time series are used throughout in order to emphasize large-scale variability while suppressing high-frequency fluctuations that are less relevant to teleconnection processes \cite{Runge2019NatCommun,runge2019detecting}.

Prior to model fitting, all variables listed in Table~\ref{tab:variables} are standardized to zero mean and unit variance. For variables with pronounced seasonality (e.g., T2m, O$_3$, and PM$_{2.5}$), the climatological monthly mean is removed before standardization to reduce seasonal leakage and to ensure that the inferred dependencies primarily reflect intermonthly variability rather than the mean annual cycle. This preprocessing strategy is kept identical across the two region--season settings to preserve comparability of the inferred causal structures.

\subsection{Lag Window Selection}

A key design choice in teleconnection-oriented temporal causal discovery is the specification of the maximum lag window $L$\cite{granger1969investigating,schreiber2000measuring,Runge2019NatCommun}. In the main experiments, we adopt

\begin{equation}
L = 12 \ \text{months},
\end{equation}

which is sufficiently long to capture seasonal memory, delayed ocean--atmosphere coupling, and the circulation adjustment timescales commonly associated with ENSO, North Atlantic variability, and mid-latitude wave propagation \cite{Yeh2018Teleconnections}.

The choice of a 12-month lag window is physically motivated. Remote SST anomalies and large-scale circulation modes may affect East Asian climate and air-quality-relevant meteorology with lead times ranging from several weeks to multiple seasons \cite{Yeh2018Teleconnections}. A substantially shorter lag window may truncate meaningful delayed pathways, whereas an excessively long lag window may increase the risk of spurious dependencies induced by low-frequency persistence \cite{Runge2019NatCommun}. To assess the robustness of the inferred structures, sensitivity experiments are further performed using alternative lag windows of $L=6$ and $L=18$ months.

\subsection{SGED-TCD Configuration}

For each target variable $X_j$, SGED-TCD learns a one-step-ahead predictor from the multivariate lagged state
\begin{equation}
Z_t=\{X_i(t-\ell)\}_{i=1,\dots,d;\,\ell=1,\dots,L},
\end{equation}
where $d$ denotes the number of variables and $L$ is the maximum lag. Unlike attention-based causal discovery frameworks that infer causal relevance from attention weights, SGED-TCD introduces explicit structural gates to represent lag-specific candidate causal relations.

\subsubsection{Variable-level temporal encoder}

We employ a variable-level temporal encoder to extract lagged representations from the monthly multivariate series. In the present study, the encoder is implemented using a grouped temporal convolutional network (TCN), so that each variable is first processed independently while delayed dependencies are captured through causal convolutions. Following the SGED-TCD design, the encoder uses grouped convolutions with the grouping factor equal to the number of variables, thereby preserving variable-wise temporal representations before cross-variable structural aggregation.

The TCN backbone is configured with two temporal blocks and two convolutional layers per block, together with causal dilated convolutions and a kernel size of 2. This design provides a compact temporal encoder while maintaining sufficient receptive field growth across lagged observations. To avoid region-specific overfitting, the same encoder configuration is used for both Eastern China and Northern China.

\subsubsection{Conditional structural gating}

The core mechanism of SGED-TCD is a conditional structural gating tensor
\begin{equation}
Z_{i,j,\tau}\in[0,1],
\end{equation}
which represents the existence strength of a candidate causal link from source variable $i$ at lag $\tau$ to target variable $j$. The gate is not interpreted as a purely dynamic attention score; instead, it is a learnable structural variable composed of a global component and a data-conditioned component:
\begin{equation}
\text{logits}_{i,j,\tau}
=
\text{base\_gate\_logits}_{i,j,\tau}
+
\text{dynamic\_logits}_{i,j,\tau}.
\end{equation}

The dynamic term is generated from pairwise source--target features, including source summaries, target summaries, multiplicative interactions, temporally aligned interaction terms, and absolute source--target differences. This design enables the model to represent both global structural tendencies and context-dependent functional interactions.

To promote sparse and interpretable graph discovery, we adopt the hard-concrete gate parameterization in the main experiments. This choice allows SGED-TCD to approximate binary structural decisions while remaining differentiable during optimization.

\subsubsection{Lag-aware target aggregation and prediction}

Given the encoded variable representations and the structural gating tensor, SGED-TCD aggregates lagged parent information for each target variable $j$ through a lag-aware structural aggregation:
\begin{equation}
m_j^t
=
\sum_{i=1}^{d}\sum_{\tau=1}^{L}
Z_{i,j,\tau}\,\phi\!\left(h_i^{t-\tau},h_j^t\right),
\end{equation}
where $h_i^{t-\tau}$ denotes the encoded representation of source variable $i$ at lag $\tau$, $h_j^t$ denotes the target representation, and $\phi(\cdot)$ is a learnable transformation. The resulting aggregated message $m_j^t$ is then fused with the target representation and mapped to the one-step prediction:
\begin{equation}
\hat{x}_j^t = g_j(m_j^t,h_j^t).
\end{equation}

In SGED-TCD, causal graph extraction is based on explicit gate values whose functional relevance is further calibrated by perturbation-based effect estimates.

\subsubsection{Training objective}

All target variables are treated as continuous monthly indices. The overall training objective of SGED-TCD combines predictive accuracy, structural sparsity, structural stability, and perturbation-effect alignment:
\begin{equation}
\mathcal{L}_{\text{total}}
=
\mathcal{L}_{\text{pred}}
+
\lambda_{\text{sparse}}\mathcal{L}_{\text{sparse}}
+
\lambda_{\text{stable}}\mathcal{L}_{\text{stable}}
+
\lambda_{\text{abl}}\mathcal{L}_{\text{abl}}.
\end{equation}

The prediction loss is defined as
\begin{equation}
\mathcal{L}_{\text{pred}}
=
\frac{1}{BT}\sum_{b,t}
\left\|
\hat{x}^{(b,t)}-x^{(b,t)}
\right\|^2,
\end{equation}
where $B$ is the batch size and $T$ is the temporal length used in optimization.

To encourage sparse graph discovery at the variable-pair level while preserving lag-specific allocation, a group sparsity term is introduced:
\begin{equation}
\mathcal{L}_{\text{sparse}}
=
\sum_{i,j}
\sqrt{
\sum_{\tau}
P(Z_{i,j,\tau}>0)^2 + \epsilon
}.
\end{equation}

To improve structural reproducibility, SGED-TCD additionally penalizes inconsistency between two structure-preserving perturbation views of the same time series. Let $S^{(A)}$ and $S^{(B)}$ denote the structural outputs under two independently perturbed views; then the stability regularization encourages
\begin{equation}
\mathcal{L}_{\text{stable}}
\propto
\left\|
S^{(A)}-S^{(B)}
\right\|^2,
\end{equation}
or an equivalent hybrid consistency criterion defined over adjacency scores.

Finally, to align structural gates with their actual functional impact, SGED-TCD introduces a perturbation-effect alignment loss. For each candidate edge $(i,j,\tau)$, the perturbation effect is quantified by the change in predictive output after ablating the corresponding lagged source information:
\begin{equation}
\delta_{i,j,\tau}
=
\mathbb{E}_t
\left[
\left|
\hat{x}_j^t(\text{original})
-
\hat{x}_j^t(\text{ablated}_{i,\tau})
\right|
\right].
\end{equation}
The alignment loss encourages consistency between the learned gate values and the perturbation-induced effects, thereby reducing the gap between structural importance and predictive relevance.

In the main experiments, model optimization uses Adam with a learning rate of $1\times10^{-3}$, a batch size of 32, and an $\ell_2$ weight decay of $1\times10^{-4}$. The loss weights are fixed at $\lambda_{\text{sparse}}=0.01$, $\lambda_{\text{stable}}=0.05$, and $\lambda_{\text{abl}}=0.1$ for both regions, and no region-specific hyperparameter tuning is performed.

\subsection{From Structural Gates to a Causal Network}

After training, SGED-TCD produces lag-specific gate values $Z_{i,j,\tau}$ and perturbation-effect estimates $\delta_{i,j,\tau}$ for all ordered source--target pairs. Rather than extracting candidate edges from attention distributions, we construct lag-specific causal scores by combining normalized gate values and normalized perturbation effects:
\begin{equation}
S_{i,j,\tau}
=
\alpha\,\bar{Z}_{i,j,\tau}
+
\beta\,\bar{\delta}_{i,j,\tau},
\end{equation}
where $\bar{Z}_{i,j,\tau}$ and $\bar{\delta}_{i,j,\tau}$ denote normalized quantities, and $(\alpha,\beta)$ are composition weights. In the main setting, we adopt the default SGED-TCD combination
\begin{equation}
\alpha=0.8,\qquad \beta=0.2.
\end{equation}

For each ordered pair $(i,j)$, the dominant lag is defined as
\begin{equation}
\tau^*_{i\rightarrow j}
=
\arg\max_{\tau\in[1,L]} S_{i,j,\tau},
\end{equation}
and the pairwise edge score is obtained by lag aggregation:
\begin{equation}
A_{i,j}
=
\max_{\tau\in[1,L]} S_{i,j,\tau}.
\end{equation}

A directed edge $i\rightarrow j$ is retained in the causal network when the following conditions are jointly satisfied:

\begin{enumerate}
    \item the aggregated edge score $A_{i,j}$ exceeds the graph-selection threshold;
    \item the dominant lag $\tau^*_{i\rightarrow j}$ remains stable under lag-window perturbation and bootstrap resampling;
    \item the edge exhibits consistent perturbation-based relevance and satisfies the robustness diagnostics described in Section~3.5.
\end{enumerate}

The resulting graph is weighted, directed, and lag-annotated. Nodes represent large-scale climate indices, regional circulation variables, boundary-layer variables, and compound-extreme indicators. Edge weights encode the relative causal importance estimated by SGED-TCD, while lag labels indicate the dominant response timescale.

\subsection{Robustness and Stability Analysis}

To evaluate the reliability of the inferred SGED-TCD networks, we perform several robustness diagnostics.

\subsubsection{Lag-window sensitivity}

Causal discovery is repeated under alternative lag windows ($L=6$ and $L=18$ months). Edges that persist across different lag configurations are regarded as more robust, whereas edges that appear only under a single lag specification are interpreted more cautiously.

\subsubsection{Block-bootstrap resampling}

Because climate time series exhibit substantial temporal autocorrelation, we apply block-bootstrap resampling using contiguous temporal segments. For each bootstrap realization, SGED-TCD is re-estimated and the inferred edges are recorded. Only edges recurring in a sufficiently large fraction of bootstrap samples are retained in the final graph.

\subsubsection{Structural consistency under perturbed views}

A distinctive feature of SGED-TCD is that structural stability is enforced not only post hoc but also during training. Two structure-preserving perturbation views are generated by injecting low-amplitude noise and applying blockwise masking to the original series. Consistency of the inferred structural gates or adjacency scores across these views is used both as a training regularizer and as an additional robustness diagnostic. Edges that remain stable across such perturbations are considered more reliable than edges that are highly sensitive to minor input changes.

\subsubsection{Ablation-based validation}

For each retained edge $i\rightarrow j$, we further examine whether ablating the lagged history of source variable $i$ around $\tau^*_{i\rightarrow j}$ produces a substantial degradation in predictive performance for target $j$. This procedure provides intervention-motivated evidence that the retained source contributes non-redundant predictive information beyond what can be explained by autocorrelation or shared drivers alone.

\section{Results and Causal Network Analysis}

\begin{table}[!htbp]
\centering
\caption{Top-K candidate causal links extracted from SGED-TCD composite
causal scores for Eastern China (warm season) and Northern China (cold
season), ranked by pre-screening edge strength.}
\label{tab:extract_topk_sidebyside}
\renewcommand{\arraystretch}{1.05}
\setlength{\tabcolsep}{3.5pt}
\scriptsize
\begin{tabular}{c l l c c c | c l l c c c}
\hline
\multicolumn{6}{c|}{\textbf{Eastern China (warm season)}} & \multicolumn{6}{c}{\textbf{Northern China (cold season)}} \\
\hline
Rank & Source $X_i$ & Target $X_j$ & $\tau^\star$ & $s_{i\to j}$ & Sign &
Rank & Source $X_i$ & Target $X_j$ & $\tau^\star$ & $s_{i\to j}$ & Sign \\
\hline
1  & T2m        & O$_3$        & 0 & 0.110 & + & 1  & PBLH     & PM$_{2.5}$ & 0 & 0.102 & -- \\
2  & Z500       & T2m          & 1 & 0.090 & + & 2  & WS10     & PM$_{2.5}$ & 0 & 0.088 & -- \\
3  & Ni\~no3.4  & Z500         & 3 & 0.081 & + & 3  & Z500     & PBLH       & 1 & 0.079 & -- \\
4  & PBLH       & O$_3$        & 0 & 0.074 & + & 4  & AO       & Z500       & 2 & 0.073 & -- \\
5  & WS10       & O$_3$        & 0 & 0.069 & --& 5  & NAO      & Z500       & 3 & 0.067 & -- \\
6  & Z500       & PBLH         & 1 & 0.063 & --& 6  & Z500     & WS10       & 1 & 0.062 & -- \\
7  & HW\_Int    & O$_3$        & 0 & 0.061 & + & 7  & HW\_Dur  & PM$_{2.5}$ & 0 & 0.058 & + \\
8  & WP\_SST    & Z500         & 3 & 0.052 & + & 8  & T2m      & HW\_Dur     & 0 & 0.052 & + \\
9  & IO\_SST    & Z500         & 2 & 0.049 & + & 9  & AO       & WS10       & 2 & 0.045 & -- \\
10 & Z500       & WS10         & 1 & 0.046 & --& 10 & NA\_SST  & Z500       & 4 & 0.041 & + \\
11 & T2m        & HW\_Int      & 0 & 0.043 & + & 11 & Z500     & T2m        & 1 & 0.039 & + \\
12 & PBLH       & HW\_Int      & 0 & 0.040 & --& 12 & T2m      & PM$_{2.5}$ & 0 & 0.034 & + \\
13 & WS10       & HW\_Int      & 0 & 0.038 & --& 13 & PBLH     & HW\_Dur     & 0 & 0.032 & -- \\
14 & Ni\~no3.4  & T2m          & 5 & 0.036 & + & 14 & WS10     & HW\_Dur     & 0 & 0.030 & -- \\
15 & WP\_SST    & T2m          & 4 & 0.033 & + & 15 & NAO      & T2m        & 4 & 0.028 & + \\
\hline
\end{tabular}
\end{table}

\begin{table}[!htbp]
\centering
\caption{Top-K robust causal links retained after SGED-TCD robustness
diagnostics for Eastern China (warm season) and Northern China (cold
season), ranked by robustness.}
\label{tab:robust_sidebyside_noR}
\renewcommand{\arraystretch}{1.05}
\setlength{\tabcolsep}{2.8pt}
\scriptsize
\begin{tabular}{c l c c c c | c l c c c c}
\hline
\multicolumn{6}{c|}{\textbf{Eastern China (warm season)}} & \multicolumn{6}{c}{\textbf{Northern China (cold season)}} \\
\hline
Rank & Edge $i\to j$ & $\tau^\star$ & $f_{\text{boot}}$ & $S_{\text{lag}}$ & $\Delta$Loss &
Rank & Edge $i\to j$ & $\tau^\star$ & $f_{\text{boot}}$ & $S_{\text{lag}}$ & $\Delta$Loss \\
\hline
1  & T2m$\to$O$_3$        & 0 & 0.91 & 1.00 & 0.15 & 1  & PBLH$\to$PM$_{2.5}$   & 0 & 0.88 & 1.00 & 0.14 \\
2  & Ni\~no3.4$\to$Z500   & 3 & 0.82 & 1.00 & 0.15 & 2  & WS10$\to$PM$_{2.5}$   & 0 & 0.80 & 1.00 & 0.12 \\
3  & Z500$\to$T2m         & 1 & 0.76 & 1.00 & 0.11 & 3  & AO$\to$Z500           & 2 & 0.74 & 1.00 & 0.10 \\
4  & WP\_SST$\to$Z500     & 3 & 0.78 & 0.67 & 0.12 & 4  & NAO$\to$Z500          & 3 & 0.71 & 1.00 & 0.09 \\
5  & Z500$\to$PBLH        & 1 & 0.74 & 1.00 & 0.08 & 5  & Z500$\to$PBLH         & 1 & 0.69 & 1.00 & 0.08 \\
6  & WS10$\to$O$_3$       & 0 & 0.70 & 1.00 & 0.08 & 6  & Z500$\to$WS10         & 1 & 0.66 & 1.00 & 0.07 \\
7  & PBLH$\to$O$_3$       & 0 & 0.68 & 1.00 & 0.07 & 7  & HW\_Dur$\to$PM$_{2.5}$ & 0 & 0.63 & 1.00 & 0.06 \\
8  & IO\_SST$\to$Z500     & 2 & 0.60 & 0.67 & 0.06 & 8  & AO$\to$WS10           & 2 & 0.60 & 0.67 & 0.06 \\
9  & Z500$\to$WS10        & 1 & 0.63 & 1.00 & 0.05 & 9  & T2m$\to$HW\_Dur        & 0 & 0.58 & 1.00 & 0.04 \\
10 & T2m$\to$HW\_Int      & 0 & 0.61 & 1.00 & 0.05 & 10 & Z500$\to$T2m          & 1 & 0.55 & 1.00 & 0.04 \\
\hline
\end{tabular}
\end{table}

\subsection{Construction of the Final Causal Networks from SGED-TCD Outputs}

We adopt a multi-stage procedure to transform the raw outputs of the
proposed SGED-TCD framework into sparse, lag-annotated causal graphs
through structural screening and robustness filtering. This procedure follows
the general logic of causal discovery in time-ordered systems and climate
applications, while replacing attention-based edge extraction with the
explicit structural-gating and effect-alignment mechanisms of SGED-TCD.

Initially, SGED-TCD is applied independently to each region--season
configuration using the standardized variables listed in Table~\ref{tab:variables}. For each
target variable, the framework learns variable-level temporal representations,
conditional structural gates over source--target--lag triplets, and one-step-ahead
predictions from lagged multivariate inputs. As described in Section~3, the
framework explicitly parameterizes each candidate relation by a gate
$Z_{i,j,\tau}\in[0,1]$, where $i$ denotes the source variable, $j$ denotes the
target variable, and $\tau$ denotes the lag. In parallel, SGED-TCD estimates
the perturbation-based functional relevance of each candidate relation by
measuring the predictive impact of ablating the lagged contribution of the
source variable around the corresponding lag.

These two outputs provide complementary evidence for graph construction.
The structural gate reflects how strongly the model tends to preserve a
candidate lag-specific dependency in the learned graph, whereas the
perturbation-based quantity reflects how much that dependency functionally
matters for prediction. Rather than relying on temporal attention distributions,
candidate causal links are therefore extracted from a composite lag-specific
causal score formed by combining normalized structural-gate values and
normalized perturbation-effect estimates:
\begin{equation}
S_{i,j,\tau}
=
\alpha\,\bar{Z}_{i,j,\tau}
+
\beta\,\bar{\delta}_{i,j,\tau},
\end{equation}
where $\bar{Z}_{i,j,\tau}$ and $\bar{\delta}_{i,j,\tau}$ denote normalized gate
and perturbation-effect quantities, respectively, and $\alpha$ and $\beta$
control their relative contributions. In the main setting, SGED-TCD adopts a
gate-dominant combination so that the structural graph remains primarily
driven by explicit gates while still being calibrated by perturbation-based
functional evidence.

For each ordered variable pair $(X_i,X_j)$, the dominant lag
$\tau^{*}_{i\rightarrow j}$ is identified as the lag that maximizes the composite
causal score:
\begin{equation}
\tau^{*}_{i\rightarrow j}
=
\arg\max_{\tau\in[1,L]} S_{i,j,\tau},
\end{equation}
and the lag-aggregated pre-screening edge strength is defined by
\begin{equation}
A_{i,j}
=
\max_{\tau\in[1,L]} S_{i,j,\tau}.
\end{equation}
This produces an initial dense directed graph in which each candidate link is
associated with both a dominant lag and a relative causal score. For
presentation purposes, we report the Top-K candidate links (Table~\ref{tab:extract_topk_sidebyside})
according to these pre-screening scores.

To suppress weak or unstable relations, candidate edges are further filtered
using a graph-selection threshold derived from the empirical distribution of
the composite edge scores. This thresholding step reduces the influence of
weak dependencies that may arise from temporal persistence, common forcing,
or limited-sample fluctuations. Unlike the original attention-based pipeline,
the present procedure evaluates candidate links using both explicit structural
evidence and perturbation-aligned predictive relevance before they are passed
to the final robustness stage.

Following the initial screening, the remaining candidate edges undergo
robustness diagnostics designed to mitigate spurious dependencies induced by
autocorrelation, persistence, and shared drivers. First, SGED-TCD is
re-estimated under alternative lag windows to examine whether the inferred
dominant lag and edge presence remain stable under moderate changes in the
maximum lag setting. Second, block-bootstrap resampling is applied to the
monthly time series so that each candidate edge can be evaluated in terms of
its recurrence frequency across bootstrap realizations. Third, ablation-based
validation is performed by suppressing the lagged history of the driver
variable around $\tau^{*}_{i\rightarrow j}$ and re-evaluating predictive
performance for the target node. Only edges demonstrating structural
stability, recurrence under bootstrap resampling, and a non-negligible
predictive effect are retained in the final graph. The Top-K robust links are
presented in Table~\ref{tab:robust_sidebyside_noR}.

For each retained edge, the final weight is normalized to represent its
relative causal importance within the network. Arrowheads indicate the
direction of influence, the color of the edge denotes the sign of the inferred causal
effect, and the dominant delay $\tau^{*}_{i\rightarrow j}$ is annotated on the edge
to indicate the characteristic timescale of the pathway. This visualization
strategy ensures that Fig.~\ref{fig:causal_network} clearly conveys the directionality, relative
importance, and temporal scale of the causal relations driven by identified teleconnection.

The final causal networks shown in Fig.~\ref{fig:causal_network} therefore represent a refined
structure obtained through explicit structural gating, perturbation-effect
alignment, and robustness-guided graph extraction. By retaining only the
strongest, most stable, and physically interpretable links, the final networks
highlight persistent pathways from remote climate modes to regional
circulation anomalies, local meteorological mediators, and compound extreme
indicators. The causal interpretation in the following subsections focuses on
these robust pathways and their relative influences.

\subsection{Overview of the Inferred Causal Networks}
\begin{figure}[!htbp]
  \centering
  \includegraphics[width=1\textwidth]{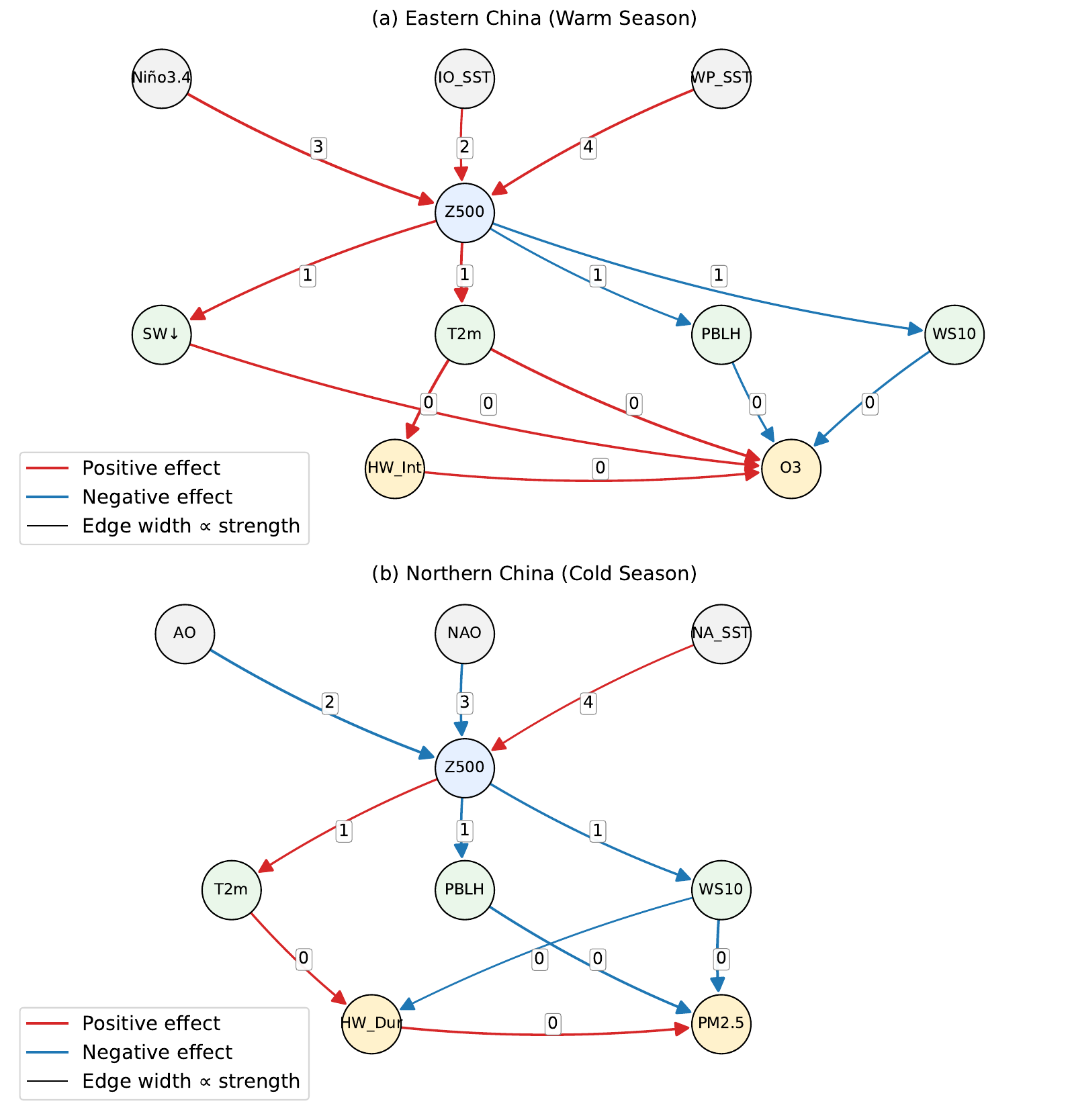}
  \caption{SGED-TCD-inferred weighted causal networks for (a) Eastern China
during the warm season and (b) Northern China during the cold season.
Nodes represent large-scale climate indices, regional circulation and
boundary-layer variables, and compound extreme indicators. Directed edges
indicate robust causal influences retained after structural screening and
robustness-guided graph extraction. Edge width is proportional to relative
causal importance, edge color denotes the sign of the effect, and numerical
labels indicate the dominant lag (months). Only edges that remain stable
under lag-window sensitivity, block-bootstrap resampling, and ablation-based
validation are retained.}
  \label{fig:causal_network}
\end{figure}

Fig.~\ref{fig:causal_network} shows the weighted causal networks inferred by the proposed
SGED-TCD framework for (a) Eastern China during the warm season and
(b) Northern China during the cold season. Nodes denote large-scale climate
indices, regional circulation variables, local meteorological mediators, and
compound-extreme indicators. Directed edges correspond to candidate causal
relations that remain after composite-score screening, lag aggregation, and
robustness-guided graph extraction.

Across both region--season settings, the inferred networks exhibit a clear
hierarchical structure. Remote teleconnection drivers primarily act on
regional circulation anomalies, which then propagate their influence to
boundary-layer and near-surface meteorological conditions before affecting
compound heatwave--air-pollution extremes\cite{WallaceGutzler1981}. This organization supports the
interpretation that the identified risks are governed by lagged, multi-step,
and region-specific causal pathways rather than by direct one-to-one forcing.

\subsection{Eastern China: Warm-Season Heatwave--Ozone Pathways}

For Eastern China during the warm season (Fig.~\ref{fig:causal_network}a), the causal network accentuates the pivotal role of the variability of the Pacific and Indian Ocean in shaping the heat wave-ozone agreement. This finding aligns with existing evidence that large-scale climate modes, including ENSO, modulate ozone pollution over China on interannual timescales via circulation anomalies \cite{Yang2022ENSO_O3, Yu2022ElNino_O3_ACP}.

Ni\~no3.4 and WP\_SST emerge as primary remote drivers, exerting delayed influences on mid-tropospheric circulation anomalies (Z500) over East Asia with lead times spanning several months. These circulation anomalies are related to a stronger or extended subtropical high in the west Pacific, which promotes subsidence, reduced cloud cover, and enhanced radiative/thermal conditions conducive to photochemical ozone production \cite{jiang2021impact,Wang2023WPSH_O3}.

Downstream of Z500, surface meteorological variables such as $T2m$, $PBLH$, and $WS10$ form a tightly interconnected subnetwork. Elevated near-surface temperatures and suppressed ventilation conditions collectively enhance ozone production efficiency and accumulation, resulting in a positive causal influence on warm-season $O_3$ concentrations \cite{Han2020LocalSynopticO3,Wang2022NCP_O3_Heat}.

Heatwave intensity ($HW\_Int$) demonstrates a causal link to both $T2m$ and $O_3$, indicating that ozone extremes preferentially coincide with periods of intensified heat stress rather than occurring as independent pollution episodes. This observation is consistent with empirical evidence of compound heatwave--ozone events and their synoptic drivers in northern/eastern China \cite{Zong2022HW_O3_Beijing,Wang2022NCP_O3_Heat}.

The identified lag structure suggests that remote SST anomalies influence ozone-related extremes on seasonal timescales, while local meteorological adjustments operate on shorter lags. This is consistent with seasonal predictability arguments based on oceanic precursors and subsequent circulation anomalies \cite{Yang2022ENSO_O3}.

\subsection{Northern China: Cold-Season Heatwave--PM$_{2.5}$ Pathways}

In contrast, the cold-season network for Northern China (Fig.~\ref{fig:causal_network}b) is predominantly governed by high-latitude circulation modes. The Arctic Oscillation (AO) and North Atlantic Oscillation (NAO) function as key upstream drivers, influencing Z500 anomalies over northern Eurasia and East Asia. This role is consistent with canonical large-scale teleconnection patterns and their established influence on cold-season climate variability \cite{ThompsonWallace2000,WallaceGutzler1981}.

Negative phases of the AO and NAO are associated with weakened mid-latitude westerlies and an enhanced meridional flow. These conditions favor circulation anomalies that promote atmospheric stagnation and persistent haze-favorable conditions over the North China Plain and the Beijing--Tianjin--Hebei region \cite{Zhang2016BTH_Haze_Circulation,Zhang2021NAO_Haze_Beijing}.

These circulation changes causally impact $PBLH$ and $WS10$, leading to shallower boundary layers and reduced pollutant dispersion. Such boundary-layer suppression is a well-documented meteorological factor controlling cold-season PM$_{2.5}$ accumulation during severe haze episodes \cite{zhong2019relationship}.

Consequently, PM$_{2.5}$ concentrations increase, particularly during cold-season heatwave-like conditions characterized by anomalously warm yet stagnant air masses. This linkage corroborates studies highlighting the role of anomalous anticyclones and large-scale precursors (including Arctic sea-ice and North Atlantic signals) in modulating cold-season haze risk in northern China \cite{Zou2017SeaIceSnowHaze,Zou2020SeaIceDeclineHaze}.

Heatwave duration ($HW\_Dur$) exhibits a stronger causal relationship with PM$_{2.5}$ than heatwave intensity, reflecting the critical role of persistent stagnation, rather than short-lived temperature peaks, in driving cold-season particulate accumulation. This duration sensitivity is congruent with evidence indicating that intraseasonal persistence of the Northeast Asian anomalous anticyclone and associated circulation regimes is crucial for sustained PM$_{2.5}$ pollution in early winter \cite{An2022NAAA_PM25}.

\subsection{Comparison of Regional Causal Structures}

An examination of the two networks reveals substantial divergences in dominant drivers, intermediate processes, and response timescales. Eastern China's warm-season extremes are primarily modulated by low-latitude SST forcing and radiative--photochemical feedbacks, whereas Northern China's cold-season extremes are more strongly influenced by high-latitude circulation variability and boundary-layer stability \cite{jiang2021impact,Han2020LocalSynopticO3,Zhang2016BTH_Haze_Circulation}.

Despite these regional distinctions, both networks share a fundamental structural characteristic: compound extremes arise via multi-step causal chains rather than direct one-to-one relationships. Remote climate modes influence regional circulation, which subsequently modulates local meteorology, thereby shaping the probability and severity of concurrent heatwave--air-pollution events. This common structural feature underscores the utility of causal network analysis for disentangling complex, multivariate dependencies that are not readily apparent from pairwise correlations alone \cite{granger1969investigating,schreiber2000measuring,Sugihara2012}.

\subsection{Implications for Compound-Extreme Attribution}

The inferred causal networks offer a coherent framework for attributing concurrent heatwave--air-pollution extremes to remote atmosphere--ocean circulation anomalies. By explicitly delineating lagged and multivariate pathways, the findings demonstrate how large-scale climate variability can precondition regional environments for compound extremes months in advance \cite{Yang2022ENSO_O3,Zou2020SeaIceDeclineHaze}.

These insights have practical implications for subseasonal-to-seasonal prediction and risk assessment, as monitoring key teleconnection indices can provide early warning signals for increased compound-extreme risk in different regions of China \cite{Yang2022ENSO_O3,Zhang2021NAO_Haze_Beijing}.

\subsection{Early-Warning Predictability of Compound Heat--Pollution Months}
\label{sec:early_warning}

Motivated by the lag-resolved causal pathways identified by SGED-TCD, we further examine whether the discovered teleconnection drivers provide actionable precursor information for early warning of compound heat--pollution months. Beyond post-hoc causal attribution, this analysis tests whether the robust upstream signals retained by the SGED-TCD framework carry practical predictive value at subseasonal-to-seasonal lead times.

We formulate an early-detection task to predict the occurrence of a compound month $Y(t)$ at lead times $k\in\{1,2,3,4\}$ months. Here, $Y(t)=1$ denotes a compound month in which both the heatwave metric and the pollutant metric exceed their respective high-quantile thresholds within the target season, whereas $Y(t)=0$ denotes a non-compound month. The prediction task is conducted at monthly resolution, consistent with the causal-discovery setting adopted throughout this study.

To isolate the contribution of teleconnection features selected by SGED-TCD, we keep the forecasting model fixed as a regularized logistic regression classifier and vary only the input feature set. Specifically, we compare five settings:
(i) \emph{Persistence}, which uses the recent history of $Y$;
(ii) \emph{Correlation-selected indices}, which retain the top-$K$ lagged teleconnection features ranked by training-period correlation;
(iii) \emph{Granger/VAR-selected indices}, which represent a linear causal baseline under the same maximum lag setting; and
(iv) \emph{SGED-TCD indices}, which include only the robust teleconnection drivers retained after structural screening and robustness diagnostics together with their dominant lags $\tau^{\ast}$.

To reduce temporal leakage, we adopt a blocked time-series split for model training and evaluation. Predictive performance is assessed using the ROC-AUC score, and uncertainty is quantified by block bootstrap resampling in order to account for serial dependence in the monthly climate--environment time series.

\begin{figure}[!htbp]
\centering
\subfloat[Eastern China (warm season): HW--O$_3$]{%
  \includegraphics[width=0.78\textwidth]{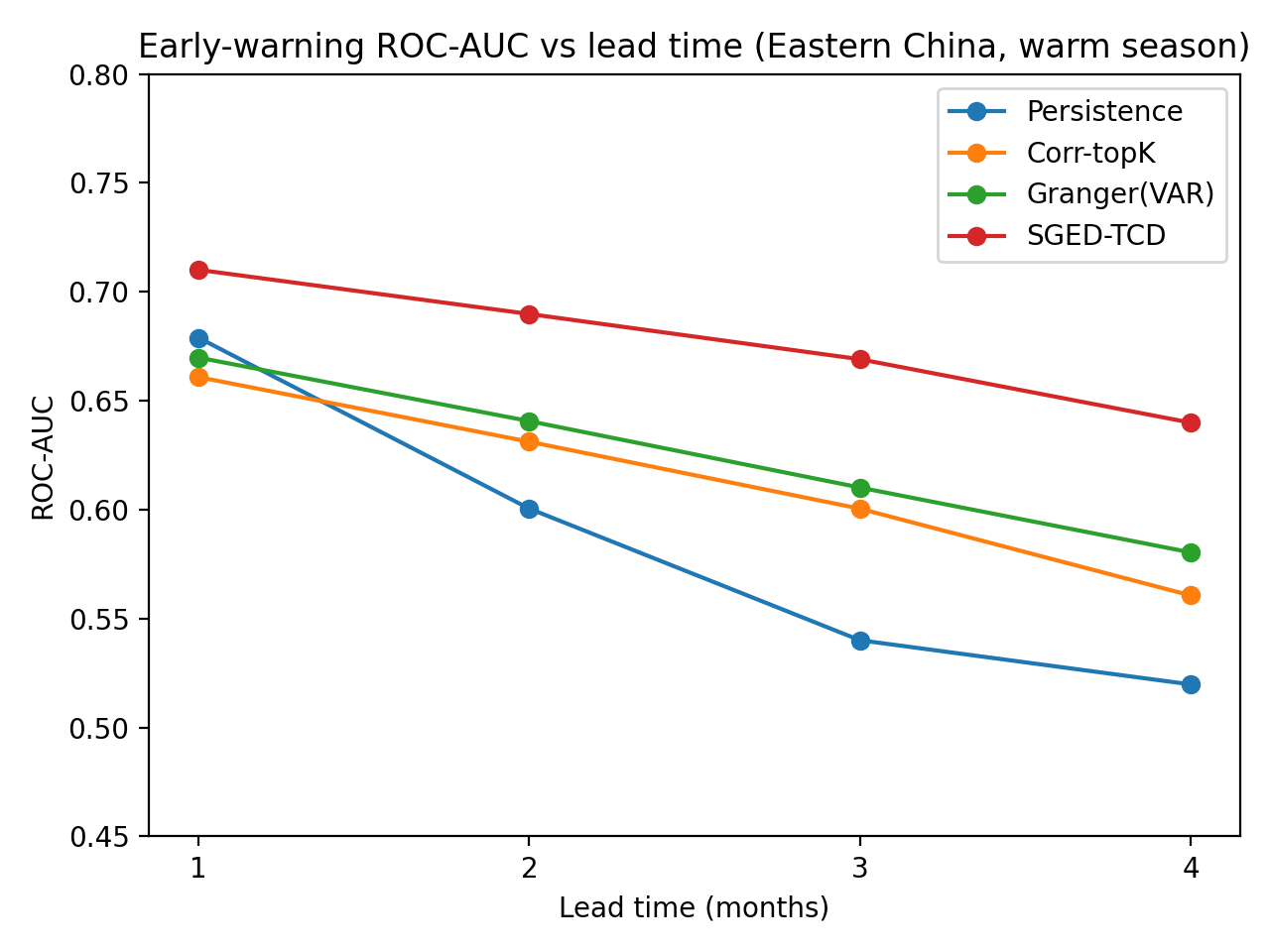}%
  \label{fig:auc_ec}}
\hfill
\subfloat[Northern China (cold season): HW--PM$_{2.5}$]{%
  \includegraphics[width=0.78\textwidth]{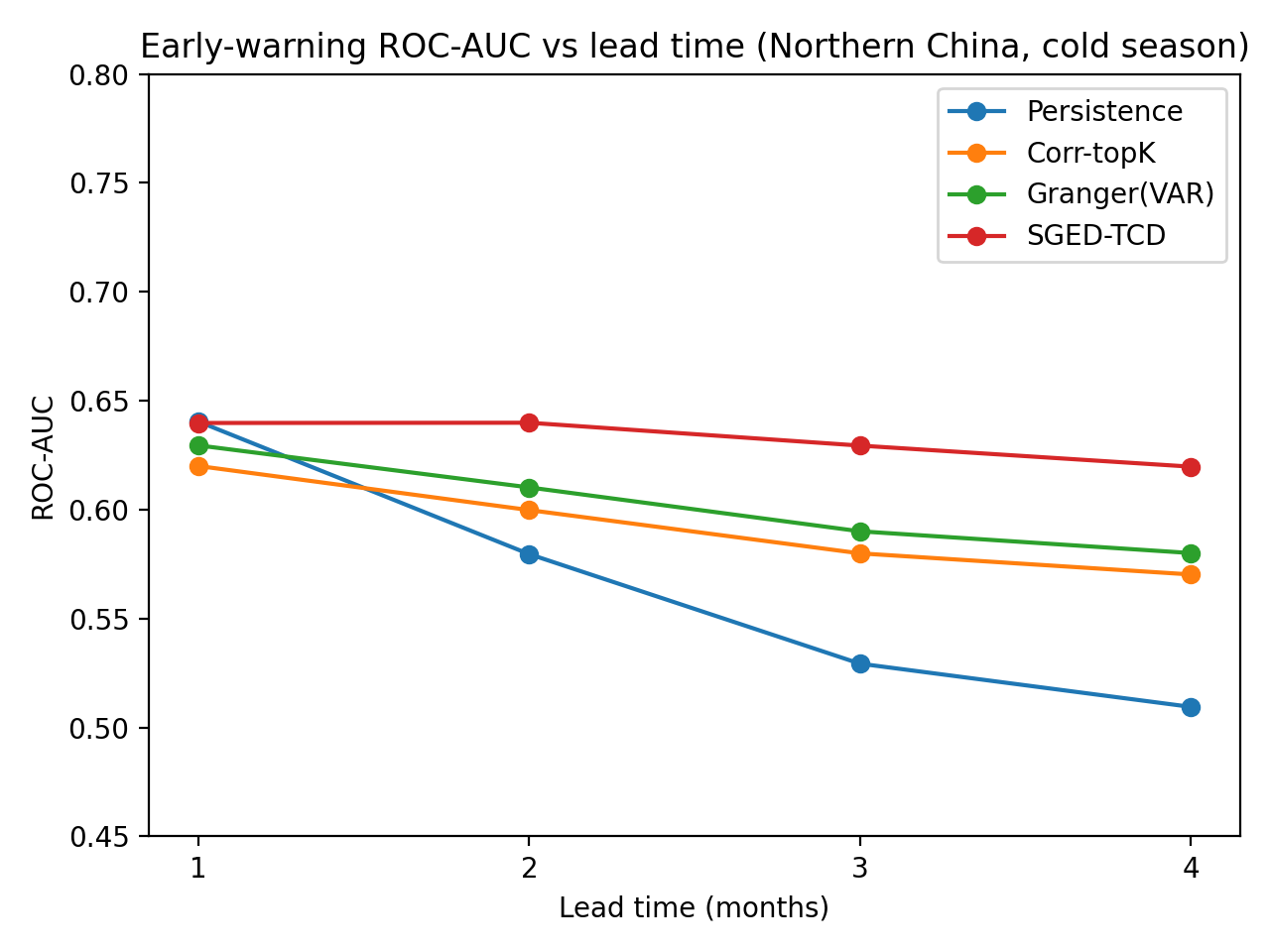}%
  \label{fig:auc_nc}}
\caption{Early-warning ROC-AUC versus lead time ($k=1$--$4$ months) for compound heat--pollution months. Teleconnection features identified by SGED-TCD provide consistently higher discrimination skill than climatology, persistence, correlation-based selection, and Granger/VAR-selected baselines, with clearer advantages at longer lead times.}
\label{fig:early_warning_auc}
\end{figure}

Fig.~\ref{fig:early_warning_auc} summarizes the early-warning ROC-AUC as a function of lead time. Across lead times, teleconnection features selected by SGED-TCD consistently improve early detection skill relative to correlation-based selection and linear Granger/VAR baselines. For warm-season compound HW--O$_3$ months over Eastern China (Fig.~\ref{fig:auc_ec}), the SGED-TCD feature set improves ROC-AUC from 0.60 to 0.67 at $k=3$ months and yields a positive Brier Skill Score of 0.065, indicating meaningful probabilistic skill beyond climatology. For cold-season compound HW--PM$_{2.5}$ months over Northern China (Fig.~\ref{fig:auc_nc}), the largest gains emerge at longer leads; for example, at $k=4$ months, SGED-TCD achieves a ROC-AUC of 0.62, compared with 0.58 for the Granger/VAR baseline. This result is consistent with the multi-month precursor pathways identified in the SGED-TCD-inferred causal network.

Overall, these results indicate that SGED-TCD is not limited to explanatory causal attribution. The robust teleconnection drivers discovered by the framework also provide practically useful lead-time information for subseasonal-to-seasonal early warning of compound heat--pollution extremes.

\section{Conclusion}

In this study, we have proposed Structural Gating and Effect-aligned Discovery for Temporal Causal Discovery (SGED-TCD), a novel and general framework for
temporal causal discovery in complex multivariate systems. The core idea of
SGED-TCD is to move beyond attention-derived graph extraction by introducing
explicit structural gates for lag-specific candidate causal relations, together
with stability regularization, perturbation-effect alignment, and combined
graph extraction. Through this design, the framework aims to improve the
structural interpretability, robustness, and functional consistency of the causal graphs inferred from time series.

Methodologically, SGED-TCD offers several key advantages. First, it uses
explicit structural gating to represent source-target-lag relations directly,
thereby reducing the ambiguity that often arises when graph structure is
inferred only indirectly from attention weights. Second, it incorporates
stability-oriented objectives so that the inferred graph remains more consistent under structure-preserving perturbations of the input series. Third, it aligns structural importance with perturbation-based predictive effects, which helps ensure that highly scored edges are not only structurally prominent but also functionally relevant. Finally, the framework integrates gate values, perturbation effects, and robustness information into a unified graph extraction strategy, yielding lag-annotated causal networks that are both sparse and interpretable.

These properties make SGED-TCD suitable for a wide range of temporal
causal discovery tasks involving nonlinearity, autocorrelation, delayed
dependencies, and multistep interactions. In particular, its general 
structure allows it to be adapted to scientific and engineering problems 
in which time-lagged interactions and interpretable causal pathways are 
of central interest.

In the present study, teleconnection-driven compound heat
wave--air-pollution extremes in eastern and northern China are used as a representative application
scenario to evaluate the effectiveness of SGED-TCD. The resulting causal
networks have shown that the proposed framework can recover hierarchical and
physically interpretable lag-resolved pathways that link remote climate drivers,
regional circulation anomalies, local meteorological mediators, and compound
extreme indicators. Beyond causal attribution, the lag-resolved teleconnection 
pathways identified by SGED-TCD provide physically interpretable precursor 
information for subseasonal-to-seasonal early warning. In practical terms,
 inferred remote drivers can be regarded as upstream risk indicators that 
signal favorable large-scale conditions for compound extremes several months 
in advance. This information can complement routine meteorological forecasting, 
air-quality prediction, and local exposure assessment by adding a causal and
teleconnection-aware source of anticipatory information.

In general, this study establishes SGED-TCD as a new and interpretable temporal
causal discovery framework. By combining explicit structural modeling,
stability-aware learning, and effect-aligned graph extraction, SGED-TCD
provides a principled basis for discovering lag-resolved causal structure in
complex time series and has the potential to support causal analysis across a
wide variety of domains beyond the representative application examined here.

\section{Acknowledgements}
This work was supported in part by China Guangxi Science and Technology Plan Project under grant AD23026096, Chile CONICYT FONDECYT Regular under grant 1181809, and Chile CONICYT FONDEF under grant ID16I10466.

\bibliographystyle{elsarticle-num}
\bibliography{references}

\end{document}